\def\year{2020}\relax
\documentclass[letterpaper]{article} 
\usepackage{aaai20}  
\usepackage{times}  
\usepackage{helvet} 
\usepackage{courier}  
\usepackage[hyphens]{url}  
\usepackage{graphicx} 
\usepackage{graphicx}  
\usepackage{times}
\usepackage{epsfig}
\usepackage{graphicx}
\usepackage{amsmath}
\usepackage{amssymb}
\usepackage[ruled]{algorithm2e}
\usepackage{enumitem}
\usepackage{bbm}
\usepackage{multirow}
\usepackage{placeins}

\nocopyright

\title{Lifelong Object Detection}
\author{
Wang Zhou, Shiyu Chang, Norma Sosa, Hendrik Hamann, David Cox \\
IBM Research\\ 
1101 Kitchawan Road, Yorktown Heights, NY 10598\\
\{wang.zhou, shiyu.chang\}@ibm.com,  \{sosa, hendrikh\}@us.ibm.com, david.d.cox@ibm.com 
}
 
\begin{document}

\maketitle

\begin{abstract}
Recent advances in object detection have benefited significantly from rapid developments in deep neural networks.  However, neural networks suffer from the well-known issue of catastrophic forgetting, which makes continual or lifelong learning problematic.   In this paper,  we leverage the fact that new training classes arrive in a sequential manner and incrementally refine the model so that it additionally detects new object classes in the absence of previous training data.  Specifically, we consider the representative object detector, Faster R-CNN, for both accurate and efficient prediction.  To prevent abrupt performance degradation due to catastrophic forgetting, we propose to apply knowledge distillation on both the region proposal network and the region classification network, to retain the detection of previously trained classes. A pseudo-positive-aware sampling strategy is also introduced for distillation sample selection. We evaluate the proposed method on PASCAL VOC 2007 and MS COCO benchmarks and show competitive mAP and 6x inference speed improvement, which makes the approach more suitable for real-time applications.   Our implementation will be publicly available.  
\end{abstract}

\section{Introduction}
\begin{figure}[t!]
    \centering
    \includegraphics[width=0.45\textwidth]{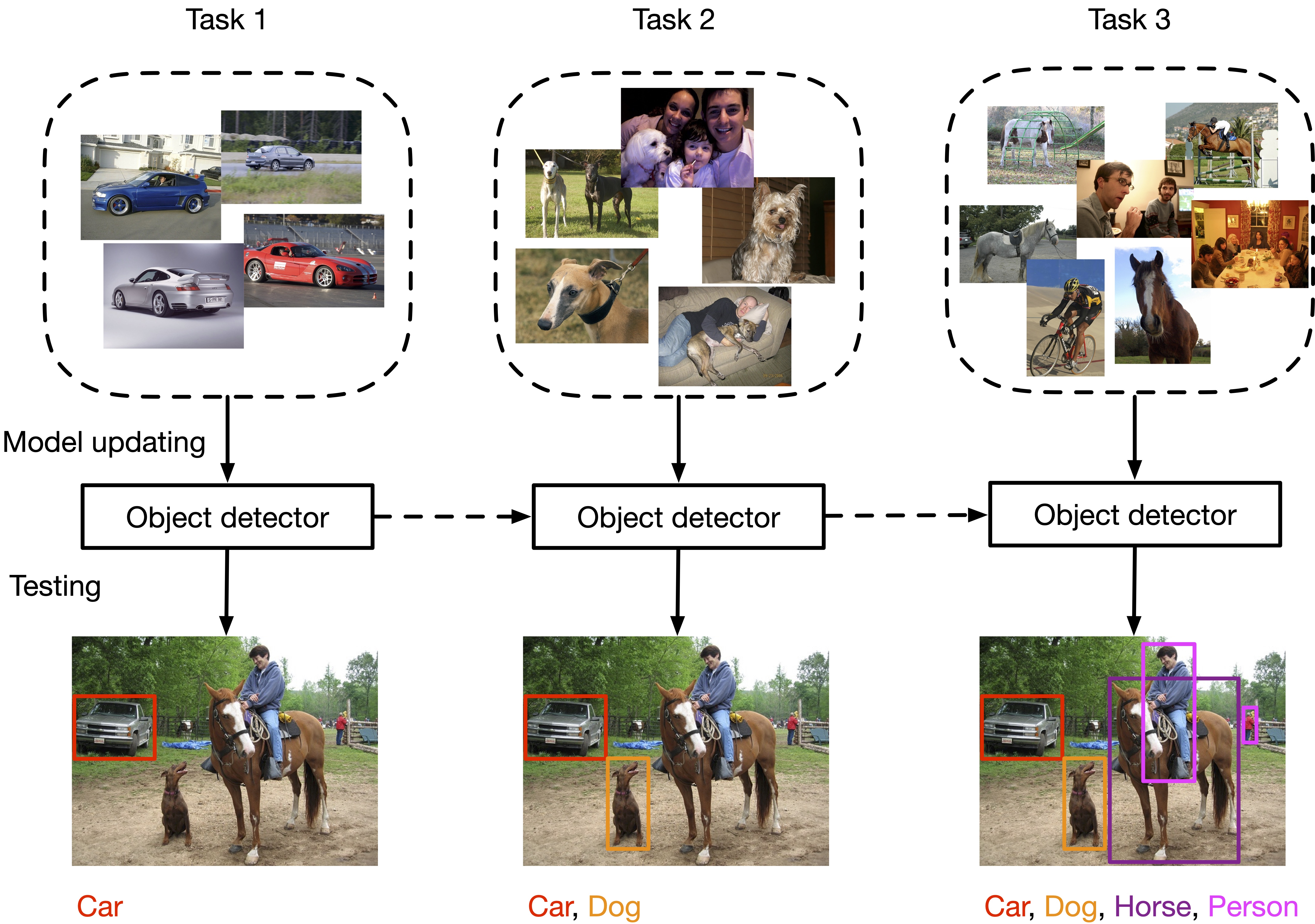}
    \caption{Illustration of lifelong object detection, where data of new object classes arrive sequentially. The goal of lifelong learning is to learn object detectors incrementally.  Specifically, the model needs to detect new object classes after re-training the detector on newly given data only (\emph{i.e.}, without accessing old training data), while not forgetting the previously trained classes.}
    \label{fig:lifelong}
    \vspace{-3mm}
\end{figure}

As the field of artificial intelligence progresses, deep learning models have matched or even surpassed human performance in a range of tasks, including image recognition \cite{he2016deep}, reading comprehension \cite{devlin2018bert}, video gaming \cite{mnih2013playing} and the game of Go \cite{silver2016mastering}.  However, despite these progresses, current neural network models are still far from ideal in a number of ways.  One key limitation of deep networks is a poor ability to learn new tasks without forgetting previously acquired knowledge.   While such lifelong learning is trivial for humans, it often represents a significant challenge for neural networks.  To alleviate the problem, in this work, we study lifelong or continual learning in the context of object detection, where new tasks/object classes are provided sequentially.   Our goal is to incrementally refine the model to provide good performance on both old and new class detection.  Figure~\ref{fig:lifelong} illustrates the process of lifelong object detection.

The phenomenon wherein neural networks forget how to solve past tasks following exposure to new tasks is known as \emph{catastrophic forgetting} \cite{mccloskey1989catastrophic,ratcliff1990connectionist}. This problem often occurs when a network is fine-tuned for a new task without considering performance degradation on the old task. There are many prior works \cite{goodfellow2013empirical,kirkpatrick2017overcoming,li2018learning,rebuffi2017icarl,van2018generative} that attempt to overcome this issue.  Most of these attempts, however, focus on the image classification problem in one form or another.  While there have been advances in continual learning in image classification, little work has been done in the context of object detection, which requires both recognition and localization of objects.  Recently, 
\cite{shmelkov2017incremental} studied the problem of incremental learning for object detection, essentially converting the detection problem to a classification problem on pre-computed region proposals with Fast R-CNN \cite{girshick2015fast}.    However, externally computed proposals are slow and inefficient for real lifelong settings, as these proposals may not cover all possible objects with high recall rates.  On the other hand, architectures such as Faster R-CNN \cite{ren2015faster} and others \cite{
dai2017deformable,
liu2016ssd,redmon2016you} are not limited to pre-computed object proposals, and it remains a challenge to incrementally learn efficient  detectors without catastrophic forgetting.  

In this paper, we propose a lifelong object detection approach that employs the Faster R-CNN architecture.  While the region proposal network (RPN) in Faster R-CNN generates class-specific proposals, adapting the network to detect new classes without forgetting is more challenging, since both the RPN and the region-based classifier R-CNN need to be updated.  To preserve the acquired knowledge for the old classes, we propose to use knowledge distillation \cite{hinton2015distilling}.  Specifically, when a new task is presented, we first retain a copy of the existing detector, which then serves as a ``teacher'' to guide the new detector.  Thus, our learning objective consists of a distillation loss (preventing forgetting old knowledge) and a supervised loss (learning new knowledge) for both RPN and R-CNN.  In addition,  since the training of Faster R-CNN relies on sampling over a large number of anchors and regions of interest (RoI),  blindly selecting them reduces the performance on old classes dramatically. We further propose a pseudo-positive-aware sampling (PPAS) strategy to facilitate retaining the performance on previously seen classes.  

To demonstrate the effectiveness and efficiency of our proposed lifelong detection method, we compare our approach with other baselines on both PASCAL VOC 2007 
and MS COCO 
benchmarks.  Extensive experimental results validate that our method delivers better detection precision, less forgetting, and higher speed efficiency.

\section{Related Work}
\label{sec:related}
Continual or lifelong learning is a long studied research problem.  The early attempts date back to \cite{ring1994continual,robins1995catastrophic
}, and it has recently received emerging attention.  In this section, we summarize some of the recent works. 

The problem considers a single model that needs to sequentially learn a series of tasks. As illustrated in \cite{van2018generative}, 
based on the availability of the task identity 
at the test time, existing lifelong learning algorithms can be roughly categorized as 1) incremental task learning, 2) incremental domain learning, and 3) incremental class learning.
The first case is the easiest lifelong learning scenario, where the task identity is always provided. 
A typical approach is to train task-specific components for each task (\emph{e.g.}, ``multi-head'' output layers) and share the rest of the network \cite{rusu2016progressive,fernando2017pathnet,mallya2018packnet,mallya2018piggyback}.
For the last two scenarios, task identity is not given during the test time.  
Incremental domain learning deals with varying input distributions but the structure of the tasks is always the same.  For example, in ``permuted MNIST'' \cite{goodfellow2013empirical}, tasks differ by permutations but all tasks need to classify the MNIST-digits.  
On the other hand, the incremental class learning is considered as the hardest and the most realistic lifelong setting, where models need to learn to recognize new classes incrementally.  An example is to classify MNIST-digits sequentially in split MNIST \cite{zenke2017continual}, where each task contains non-overlapping digits. 

Compared to the incremental task learning, models for the last two settings are vulnerable to catastrophic forgetting.  Mitigating the forgetting has been addressed in many previous studies, which can be viewed as 1) approaches using regularized optimization and 2) approaches modifying training data.  

The intuition of regularization-based approaches is that instead of optimizing the full neural networks on every task,  the parameters that are important to solve past tasks are penalized to change.  These methods have been proven effective for over-parameterized models.  EWC \cite{kirkpatrick2017overcoming
} quantifies the importance of weights to previous tasks and selectively alters the learning rates of weights using the Fisher information.   
Similarly, Synaptic Intelligence (SI) \cite{zenke2017continual} measures the synapse consolidation strength in an online fashion as a regularization.  In addition, 
RWalk \cite{chaudhry2018riemannian}  introduce two metrics to quantify forgetting and intransigence in continual learning, 
which is a generalization of EWC and SI with a theoretically grounded KL-divergence based perspective.

The second type of approach is to augment the training data for each new task with ``pseudo examples'', which characterize the data distribution of previous tasks.   The simplest option is to store examples from previous tasks and replay these data when new tasks arrive.  Representative works include iCaRL \cite{rebuffi2017icarl}, variational continual learning (VCL) \cite{nguyen2017variational}, gradient episodic memory (GEM) \cite{lopez2017gradient}, 
end-to-end incremental learning \cite{castro2018end}, \emph{etc}.  However, due to data privacy concerns, these methods are not always applicable.  Another option is to label the input data of the current task with the model trained from previous tasks.  Learning without forgetting \cite{li2018learning} uses knowledge distillation \cite{hinton2015distilling} in combination with standard cross-entropy loss.  Other approaches using distillation loss include the encoder-based method \cite{rannen2017encoder} and incremental moment matching \cite{lee2017overcoming}.  Some of the recent papers also consider generating data from previous tasks using a deep generative model \cite{lesort2018generative,
shin2017continual,van2018generative
}.   
Then, the model for the main task can be updated in a multi-task learning fashion using both the generated data and the data of the new task.  

Although there have been many works addressing the problem of lifelong learning, they all study the classic classification problem.  In this work, we consider the most realistic lifelong setting (the incremental class learning) in the context of object detection.  The only work that we are aware of that address the same problem is the one proposed by 
\cite{shmelkov2017incremental}.  Our method is significantly different from Shmelkov \emph{et al.} 
since we consider a more challenging setup where there are no pre-defined proposals (\emph{i.e.} EdgeBox \cite{zitnick2014edge}) provided.   Moreover, this also makes our approach more suitable for real lifelong setting since we do not assume the newly given objects can be detected by the external proposals.

\section{Lifelong Object Detection}\label{sec:method}

\begin{figure*}[!t]
    \centering
    \includegraphics[trim={0 6.5cm 3.0cm 0},clip,width=0.8\textwidth]{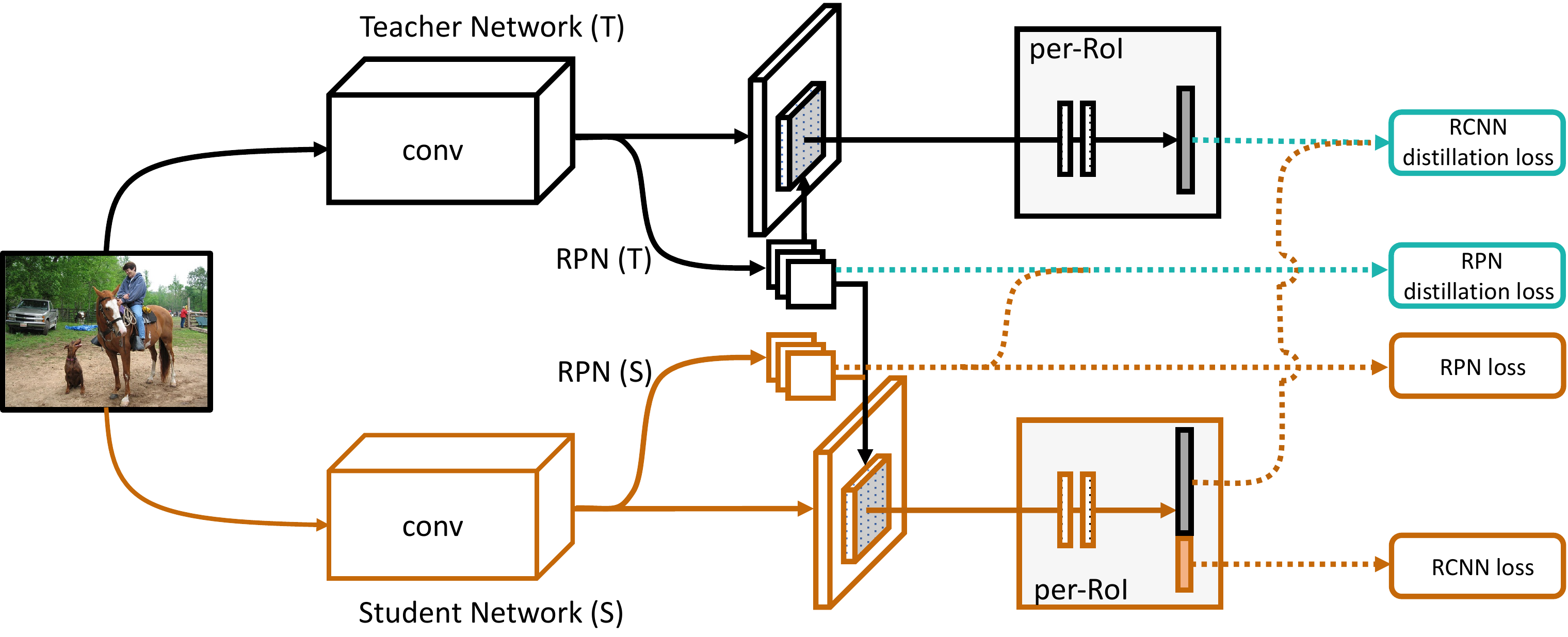} 
    \caption{The architecture of lifelong object detection. A student network $\mathbf{S}$ is a modified copy of the teacher network $\mathbf{T}$, with the number of output channels increased to accommodate the new classes.  During training, an input image is fed to both the teacher and the student network.  
    The output of RPN and R-CNN of $\mathbf{T}$ is used to ``guide'' the training of $\mathbf{S}$ via distillation losses.  A PPAS strategy is proposed to sample anchors and RoIs used to compute the RPN and R-CNN distillation losses. The student network is trained via minimizing both the supervised losses to learn the new classes and the distillation losses to preserve old knowledge. }
    \label{fig:network}
\end{figure*}

In this section, we detail our method for the end-to-end lifelong object detection using the Faster R-CNN structure \cite{ren2015faster}.   The overall architecture of our model is shown in Figure \ref{fig:network}.   At a high level, to learn a new task in the class incremental setting, we employ the teacher-student scheme \cite{hinton2015distilling}, where the teacher network ($\mathbf{T}$) is a frozen copy of the detector trained on existing classes, and the student network ($\mathbf{S}$) is a modified network with an increased number of output channels to accommodate newly added classes.  The student imitates the teacher network in order to keep the ability to detect old object classes but also learns to detect new classes it is being trained on.  This is achieved via a combination of distillation loss for old classes and supervised detection loss for new classes.   Since both RPN and R-CNN need to be modified to incorporate the new object detection, this is a much more challenging task.  We first describe the network architecture, 
and then we explain the distillation loss for both RPN and R-CNN sub-networks. 
Lastly, we propose an important sampling strategy for network distillation, \emph{i.e.}, how to select anchors for RPN distillation and RoIs for R-CNN distillation.

\subsection{Base network structure}\label{sec:frcnn}


The details of our Faster R-CNN implementation are as follows.  A ResNet-50 \cite{he2016deep} backbone pre-trained on ImageNet is used as the feature extractor for fair comparison with \cite{shmelkov2017incremental}.  Following the setup in \cite{he2016deep,chen2017implementation}, the RPN module is 
inserted before the last stride convolution layers of ResNet-50. The RPN generated proposals are used to perform RoI pooling on the feature map. 
Those features corresponding to relevant objects are converted into fixed-length vectors, which are fed into the R-CNN.
Then R-CNN tries to accomplish two goals
: 1) region classification (each channel of the softmax output represents the probability of the class, including the ``background'' class), and 2) per-class bounding box refinement, respectively. 

Rather than training the RPN and R-CNN via a four-step alternating optimization as proposed in \cite{ren2015faster}, we jointly train the entire network with a multi-task loss.  For RPN, a mini-batch of 256 anchors is randomly sampled from all anchors with the ratio of positive and negative anchors up to 1:1. The loss function for RPN is defined as:  
\begin{equation}
\label{eq:rpn_sup}
\small
\mathcal{L}^{\text{RPN}} = \mathcal{L}^{\text{RPN}}_{\text{cls}} + \lambda~ \mathcal{L}^{\text{RPN}}_{\text{reg}},
\end{equation}
where the classification loss $\mathcal{L}^{\text{RPN}}_{\text{cls}}$ is a binary cross-entropy loss over two channels (\emph{i.e.}, background vs object), and the regression loss $\mathcal{L}^{\text{RPN}}_\text{{reg}}$ is a robust smooth $\ell_1$ loss defined in \cite{girshick2015fast} applied on the positive anchors only.  

After being adjusted with the proposal refinement output, all anchors are then ranked by the objectness score, filtered by non-maximum suppression (NMS), and selected to keep the top 2,000 proposals for the R-CNN training.  Note that during inference, typically only the top 300 proposals are selected and therefore fewer regions need to be processed by the R-CNN network, leading to faster detection.  These proposals are further sampled randomly to form a mini-batch of 128 RoIs, with a positive to negative ratio up to 1:3.  The loss of R-CNN detection is defined as:
\begin{equation}
\label{eq:rcnn_sup}
\small
\mathcal{L}^{\text{RCNN}} = \mathcal{L}^{\text{RCNN}}_{\text{cls}} + \lambda'~ \mathcal{L}^{\text{RCNN}}_{\text{reg}}.
\end{equation}
Similarly, the classification loss $\mathcal{L}^{\text{RCNN}}_{\text{cls}}$ is a cross-entropy over all object classes including background, and the bounding box regression loss $\mathcal{L}^{\text{RCNN}}_{\text{reg}}$ is a robust smooth $\ell_1$ loss, but only for positive RoIs.   The total supervised loss is a combination of $\mathcal{L}^{\text{RPN}}$  and $\mathcal{L}^{\text{RCNN}}$.   More details on Faster R-CNN training and  sample selection can be found in \cite{girshick2015fast,ren2015faster}.

\subsection{Teacher-student network training}
\label{sec:distil_loss}
To detect new objects without forgetting, our lifelong detection approach requires the use of outputs from a teacher network to ``supervise'' the training of a new student network via distillation losses for both the RPN and the R-CNN.  As shown in Figure~\ref{fig:network}, before learning to incorporate new classes to the model, we first keep a fixed copy of the model that detects the original set of objects $C_{\text{old}}$, and use it as the teacher network $\mathbf{T}$.  Then, we train a student network $\mathbf{S}$ on the data of new object classes $C_{\text{new}}$ so that it can detect both the old and the new classes. 

Specifically, the student network is initialized with the parameters of the teacher network $\mathbf{T}$.  The number of the R-CNN classification and regression outputs are then increased by $\left|C_{\text{new}}\right|$ and $4\times\left|C_{\text{new}}\right|$ accordingly.   Unlike \cite{shmelkov2017incremental} that adds a new fully-connected layer per task increment,  we directly extend the size of our output channels.  These newly added weight matrices are initialized randomly.  In this way, the student network has exactly the same structure after adding new classes as if it were trained from scratch.  In addition, since the original RPN of the teacher network possibly treats the new classes $C_{\text{new}}$ as background, directly adopting the RPN from the previous network without modification will result in few proposals covering the new classes.  Thus,  not only the R-CNN but also the RPN network need to be updated.  Next, we will detail the learning strategies for both modules.  

The input images of $C_{\text{new}}$ are first fed to the teacher network $\mathbf{T}$.  We gather the network outputs for distillation, which includes RPN outputs, selected top 2,000 RoI proposals, and R-CNN outputs of these selected proposals\footnote{Given that the teacher network is not updated during training, these outputs can be cached to 
speed up the training.}. All of these outputs will be used as candidates to distill the teacher network $\mathbf{T}$.   Then, these input images of $C_{\text{new}}$ are fed to the student network $\mathbf{S}$.  Apart from the supervised learning losses for the new classes $C_{\text{new}}$ defined in Equation~(\ref{eq:rpn_sup}) and (\ref{eq:rcnn_sup}) (depicted as the orange modules in Figure~\ref{fig:network}), distillation losses are designed for both the RPN and R-CNN sub-networks (the green modules in Figure~\ref{fig:network}). To distill the RPN, we compare the classification and regression outputs of RPN from both $\mathbf{T}$ and $\mathbf{S}$ at the same $N$ carefully selected anchors.  The pairwise RPN distillation loss is
\begin{equation}
\label{eq:rpn_dis}
\small
\mathcal{L}^{\text{RPN}}_{\text{dist}} = \frac{1}{N}\sum_i f_{\text{cls}}(p_i^{\mathbf{S}}, p^\mathbf{T}_i) + \mathbbm{1}[p^\mathbf{T}_i \ge 0.5] \cdot f_{\text{reg}}(t_i^{\mathbf{S}}, t_i^\mathbf{T}),
\end{equation}
where $p^{\mathbf{S}}$ and $p^\mathbf{T}$ are the softmax outputs of classification layers for both student and teacher RPN while $t^{\mathbf{S}}$ and $t^\mathbf{T}$ are the proposal refinement outputs.  $f_{\text{cls}}$ and $f_{\text{reg}}$ are the cross-entropy and smooth $\ell_1$ loss, respectively. 
The indicator function $\mathbbm{1}(\cdot)$ evaluates to 1 when the objectness score $p^\mathbf{T}_i \ge 0.5$, and 0 otherwise.  Here, 256 anchors are selected ($N=256$). 

On the other hand, for the R-CNN distillation, we first select $M$ proposals from the 2,000 candidates generated by the teacher RPN.  Then we pass these selected proposals through the student R-CNN to compute the detection outputs.  We set $M=128$ throughout this paper.  The R-CNN distillation loss is calculated as
\begin{equation}
\label{eq:rcnn_dis}
\small
\mathcal{L}^{\text{RCNN}}_{\text{dist}} = \frac{1}{M}\sum_i g_{\text{cls}}(q_i^{\mathbf{S}}, q^\mathbf{T}_i) + \mathbbm{1}[z^\mathbf{T}_i \ge 0.5] f_{\text{reg}} (\tau_i^{\mathbf{S}}, \tau_i^\mathbf{T}),
\end{equation}
where $g_{\text{cls}}$ is the $\ell_2$ loss.  $q_i^{\mathbf{S}}$ and $q_i^{\mathbf{T}}$ denote the mean-subtracted logits for classes of $C_{\text{old}}$ only, obtained before the softmax layers of the student and teacher R-CNN, respectively \cite{shmelkov2017incremental}, while $\tau^{\mathbf{S}}$ and $\tau^\mathbf{T}$ are the bounding box regression outputs.  The regression loss is similar to the one in Equation (\ref{eq:rpn_dis}), but with a different indicator function.  $z^\mathbf{T}_i$ is the maximum probability of old classes (excluding the background class) of the softmax layer in the teacher's R-CNN.   In other words, a bounding box needs to be predicted as an object with a probability more than 0.5 by the teacher network in order to be considered in the regression loss.  

The overall training objective for lifelong detection combines the supervised loss for the new classes and distillation loss for the old classes, which is given as
\begin{equation}
\label{eq:total_loss}
\small
\mathcal{L}_{\text{total}} = \mathcal{L}^{\text{RPN}} + \lambda_1 ~\mathcal{L}^{\text{RCNN}} + \lambda_2 ~\mathcal{L}^{\text{RPN}}_{\text{dist}} + \lambda_3~ \mathcal{L}^{\text{RCNN}}_{\text{dist}}.
\end{equation}
The hyper-parameters $\lambda_1$, $\lambda_2$ and $\lambda_3$ are set to 1.0 to balance the supervised and distillation loss of each component.

\subsection{Pseudo-Positive-Aware Sampling}\label{sec:sampling}
The key idea behind the teacher-student network training is to make the predictions of the student network similar to those of the teacher network for the old classes.  However, there are an excessive number of anchors and RoIs to choose from the teacher network $\mathbf{T}$ for the purpose of distillation. It is important to sample those that help to keep the detection performance on old classes while not sabotaging the training of new classes.   Hence, we propose the PPAS strategy.  

For a given input image, if an output bounding box of $\mathbf{T}$ has a probability score greater than 0.5 for any class other than background, we consider it as a ``pseudo-positive box''.   Based on this definition, we then define the ``pseudo-positive anchor'' and ``pseudo-positive RoI'' of the student network.   Following the definition of positive and negative anchors \cite{ren2015faster}, an anchor in the student RPN is recognized as a pseudo-positive or pseudo-negative if the intersection over union (IoU) between the anchor and any pseudo-positive box is greater than 0.7 or less than 0.3, respectively.  Similarly, the pseudo-positive or pseudo-negative RoIs of $\mathbf{S}$ are the ones that have IoU greater than 0.5 or between 0.1 and 0.5 \cite{girshick2015fast} with the pseudo-positive boxes.

One problem that causes the newly updated model to forget the knowledge of previous classes is that old object classes in the newly given images, if there is any, are treated as ``background''.  This is because that the labeled bounding boxes in the new data are only the ones belonging to the new classes.  To avoid this issue,  we need to be aware of the pseudo-positive boxes when sampling negative anchors and RoIs for training the student detector on the new classes. Specifically, the proposed PPAS first
\begin{enumerate}[noitemsep,nolistsep]
\item excludes pseudo-positive anchors from being selected as negative anchors; and
\item excludes pseudo-positive RoIs from being selected as negative RoI samples.
\end{enumerate}
This ensures that we prevent the old objects from being treated as background.
After the filtering, the PPAS samples 256 anchors out of 512 with the highest objectness scores in $\mathbf{T}$ for RPN distillation, and 128 RoI proposals out of 256 that have the lowest background scores for R-CNN distillation.  We conduct ablation studies for the proposed PPAS 
and show it helps to preserve the detection of old classes. We summarize the pipeline of our lifelong detection in Algorithm~\ref{alg}. 

\LinesNumbered
\begin{algorithm}[t]
\small
\label{alg}
\caption{Lifelong Object Detection}
\SetAlgoLined
\textbf{Require:} The current detector $\mathbf{T}$, and the new task dataset $\mathcal{D}$ for new object classes $C_{\text{new}}$ \\
\textbf{Initialization:}
Initialize the parameters of a student model $\mathbf{S}$: $\theta^{\mathbf{S}} \leftarrow \theta^{\mathbf{T}}$. Add additional $|C_{\text{new}}|$ number of output channels to the student's R-CNN. \\
\textbf{Training on the new task:} \\
\While{\emph{not done}}{
    \For{\emph{labeled} $x \in \mathcal{D}$}{
      Forward pass on $\mathbf{T}$: $\{p^{\mathbf{T}}, t^{\mathbf{T}}, q^{\mathbf{T}}, \tau^{\mathbf{T}}\} \leftarrow \mathbf{T}\left(x\right)$ \\
      Obtain the RPN outputs of $\mathbf{S}$: $\{p^{\mathbf{S}}, t^{\mathbf{S}}\} \leftarrow \mathbf{S}\left(x\right)$ \\
      $\mathcal{L}^{\text{RPN}}$, $\mathcal{L}^{\text{RPN}}_{\text{dist}} \leftarrow$ Compute Eq.~(\ref{eq:rpn_sup}) and (\ref{eq:rpn_dis}) with PPAS \\
      Sample 128 RoIs out of 2,000 from $\mathbf{S}$ via PPAS \\
      Sample 128 RoIs out of 2,000 from $\mathbf{T}$ via PPAS \\
      Obtain the R-CNN outputs of $\mathbf{S}$ on sampled RoIs \\
      $\mathcal{L}^{\text{RCNN}}$, $\mathcal{L}^{\text{RCNN}}_{\text{dist}} \leftarrow$ Compute Eq.~(\ref{eq:rcnn_sup}) and Eq.~(\ref{eq:rcnn_dis}) \\
      $\mathcal{L}_{\text{total}} \leftarrow$ Compute Eq.~(\ref{eq:total_loss}) \\
      Update $\mathbf{S}$: $\theta^{\mathbf{S}} \leftarrow \theta^{\mathbf{S}} - \alpha \nabla_{\theta^{\mathbf{S}}}\mathcal{L}_{\text{total}}$
      }
  }
\textbf{Update current detector:}  $\mathbf{T} \leftarrow \mathbf{S}$ 
\end{algorithm}

\section{Experiment}
\label{sec:experiment}

\subsection{Experiment settings}

We perform our experiments on both the PASCAL VOC 2007 detection benchmark 
and the Microsoft COCO challenge dataset. 
VOC 2007 consists of about 5K training and validation images and 5K test images over 20 object categories.  On the other hand, COCO has 80K training images and 40K validation images covering 80 object classes.  For VOC, we follow the convention setup and report the standard mean average precision (mAP) at 0.5 IoU threshold. 
For COCO, we train on the training set and evaluate on the first 5K images of the validation set (minival).  Additional average mAP over IoU from 0.5 to 0.95 is also reported, which is commonly used in COCO detection challenges.

To seek a fair comparison, we follow the same experiment settings as reported in \cite{shmelkov2017incremental}, which are as follows.  We start with selecting a subset of classes from the training dataset as the old class set $C_{\text{old}}$.  A Faster R-CNN model is then trained only on the data containing classes in $C_{\text{old}}$, which is later used as the teacher network $\mathbf{T}$.  The rest of the classes are treated as the new class set $C_{\text{new}}$ in either a single new task or multiple sequential tasks.  In both settings, labeled bounding boxes that do not belong to the current task are not provided, and images with no objects of the current task are excluded.   

We filter the final detection outputs that are less than 0.5, although typically a higher recall and mAP can be achieved without filtering.  Following \cite{castro2018end}, we repeat each experiment five times and report the average results to mitigate the randomness in training VOC.  We use the code\footnote{https://github.com/kshmelkov/incremental\_detectors} provided by \cite{shmelkov2017incremental} to run the same experiments and use it as a baseline\footnote{We observe there are small performance differences between some of our reproduced results and the ones reported in the original paper. We report the best performance we can reproduce with the provided code. }, which is denoted as ``Seq-Fast''.  Our lifelong detection is implemented with Tensorflow \cite{
chen2017implementation}
, and implementation details can be found in Appendix.

\begin{table}[!t]
\small
\centering
\caption{Test results of ``19+1'' on VOC 2007 dataset. We consider both fine-tuning only (``FT'') and distillation methods (``Distil'').}
\label{tab:19plus1}
\begin{tabular}{l|l|ccc}
\multicolumn{2}{l|}{} & Old & New & All \\ \hline \hline
\multicolumn{2}{l|}{$\mathbf{T}$(1-19)} & 70.6 & - & - \\ \hline
\multicolumn{1}{c|}{\multirow{3}{*}{\rotatebox{90}{FT}}} & Entire & 19.7 & 30.5 & 20.3 \\
\multicolumn{1}{c|}{} & Fix RPN & 33.1 & 33.0 & 33.1 \\
\multicolumn{1}{c|}{} & Fix RPN and Conv & 40.2 & 19.7 & 39.2 \\ \hline
\multirow{3}{*}{\rotatebox{90}{Distil}} 
 & Seq-Fast \shortcite{shmelkov2017incremental} & \multicolumn{1}{l}{67.9} & \multicolumn{1}{l}{54.6} & \multicolumn{1}{l}{67.2} \\
 & Ours w/o PPAS & 69.4 & 49.5 & 68.4 \\
 & Ours w/ PPAS & 70.5 & 53.0 & 69.6 \\ \hline
\multicolumn{2}{l|}{$\mathbf{T}$(1-20)} & 70.4 & 66.6 & 70.2
\end{tabular}
\end{table}

\subsection{Adding one class}

The simplest setting of sequential learning of new classes on top of an existing object detector is to add just one class.   We take the first 19 classes in alphabetical order from VOC dataset as $C_{\text{old}}$ and train a Faster R-CNN denoted as $\mathbf{T}$(1-19).   The last class, ``TV monitor'', is then used as $C_{\text{new}}$.  This simple experiment setup is denoted as ``19+1'', where the first number is $|C_{\text{old}}|$ and the latter one is $|C_{\text{new}}|$.

Table~\ref{tab:19plus1} shows the evaluation results of ``19+1'' with different baselines.  Specifically, we consider both fine-tuning only and distillation based methods.  We report the mAP for old, new and all classes, which are listed in three columns in the table.  If we fine-tune the entire network without considering the detection of the old classes, the performance on $C_{\text{old}}$ slumps from 70.6\% of $\mathbf{T}$(1-19) to 19.7\%, manifesting the problem of catastrophic forgetting.  In addition, the performance on $C_{\text{new}}$ is 30.5\%, which is far below the upper-bound 66.6\% denoted as $\mathbf{T}$(1-20) that trains all classes together.  This is potentially caused by overfitting since there are only 256 training images of ``TV monitor''. 
We also consider cases that fix different modules of the Faster R-CNN while fine-tuning on the task.  For these restricted fine-tuning cases, the performance on the old classes is better retained, but still much worse than other distillation based methods.  

\begin{figure*}[!th]
    \centering
    \includegraphics[trim=0cm 0cm 0cm 0cm, width=1.\textwidth]{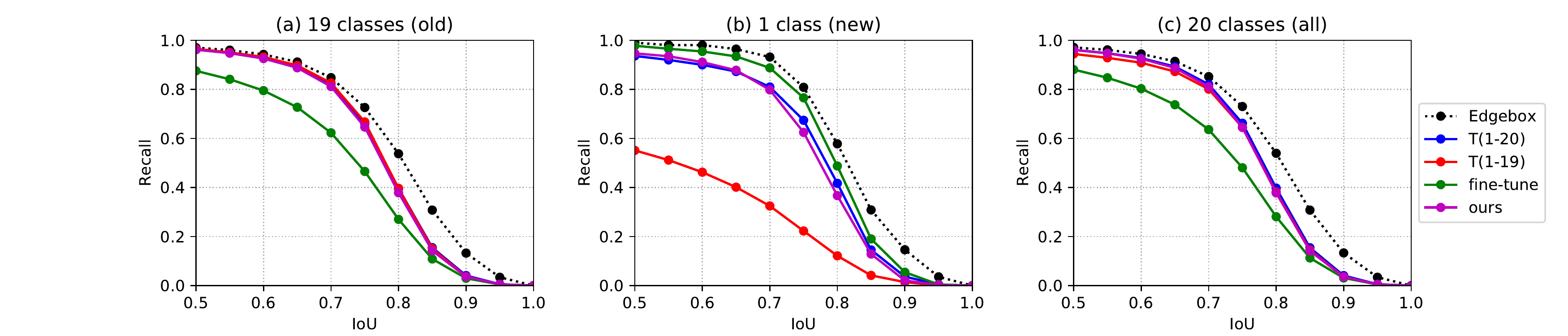}
    \caption{Comparison of object proposal recalls for the ``19+1'' setup. Recalls of the  RPN of different Faster R-CNN methods are plotted as solid lines, and the  Edgebox is plotted as a dotted line. Note that recalls of RPN are calculated based on only \textbf{300} proposals, while recall of Edgebox is calculated from \textbf{2,000} proposals. From left to right, the panels show the average recalls for the old classes, the new class (``TV monitor''), and all 20 classes.}
    \label{fig:recalls}
\end{figure*}

\begin{table*}[!h]
\small
\centering
\caption{Test results of adding new classes of VOC 2007 sequentially. Reported results are the ones when all tasks are added.  Results in each column are shown in the format of ``Old / New / All''.}
\vspace{1mm}
\label{tab:sequential}
\begin{tabular}{l|c c c c}
          & Fine-Tune & Seq-Fast \shortcite{shmelkov2017incremental} & Ours w/o PPAS & Ours w/ PPAS \\
\hline \hline
``10+1...'' & 3.2 / 2.1 / 2.7 & 53.5 / 38.7 / 46.1 & 22.9 / 17.7 / 20.3 & 54.2 / 38.2 / 46.2 \\
``10+2...'' & 8.4 / 8.4 / 8.4 &  55.2 / 46.3 / 50.8 & 35.4 / 27.9 / 31.7 & 55.5 / 44.5 / 50.0 \\
``10+5+5'' & 13.7 / 22.5 / 18.1 & 58.7 / 51.7 / 55.2 & 48.9 / 48.1 / 48.5 & 60.3 / 53.1 / 56.7\\
``10+10'' & 12.3 / 60.4 / 36.4 & 62.8 / 58.9 / 60.8  & 58.7 / 60.9 / 59.8 & 63.5 / 60.0 / 61.8 \\
\end{tabular}
\end{table*}

Applying distillation and PPAS strategy (``Ours w/ PPAS''), the detection of the old classes is well maintained (70.5\% vs 70.6\%), and the performance on the new class is boosted, outperforming other baselines.  We also notice that if we replace the proposed PPAS with random sampling (``Ours w/o PPAS''), the performances on both the new and the old classes are reduced.   We also repeat the ``19+1'' experiment taking each of the 20 classes in VOC dataset as the newly added class. The average mAP for all test cases is 68.7\% and the standard deviation is 0.9\%, which is very competitive to the upper bound of 70.2\%.   The full list of the results is shown in Appendix. 

In order to examine whether our lifelong learning method helps to adapt the RPN network to generate proposals for the new class with less degradation on the old classes, we compare the proposal recalls before and after adding the new class in Figure~\ref{fig:recalls}.  The recall of ``TV Monitor'' is far from ideal if we directly apply network $\mathbf{T}$ on the new class images, which is shown as the red line in Figure~\ref{fig:recalls}(b).  If we fine-tune the network (green lines), the recall of the new class surpasses the upper bound $\mathbf{T}$(1-20) (blue lines), however, the average recall of the old 19 classes drops substantially and so does the overall average recall for 20 classes, which are shown in Figure~\ref{fig:recalls}(a) and (c), respectively.  In general, a higher recall of the object proposals helps to detect objects better. This is because the subsequent region classification relies on the pooled features within these proposals.  Thus, the mAP of old classes for the fine-tuning baseline is largely reduced.  Compared to the upper bound model, our approach (purple lines) improves the recall for the new class while remaining a similar recall for the old 19 classes.  Furthermore, compared to the recall of Edgebox\cite{zitnick2014edge} which is computed from 2,000 proposals in Figure~\ref{fig:recalls}, the recall of the RPN is comparable but the RPN requires only 300 proposals. The reduction in the number of required proposals leads to faster detection, as illustrated in later section. 


\subsection{Adding multiple classes at once}\label{sec:once}


We also study the case where the new task contains multiple object classes.  Specifically we consider a ``10+10'' setting, where the first 10 classes of VOC are used as $C_{\text{old}}$, and the rest are $C_{\text{new}}$.  As shown in the last row of Table~\ref{tab:sequential}, by simply fine-tuning the entire network, the mAP on the new task achieves 60.4\%.  Compared to the ``19+1'' case, fine-tuning performs much better on the new classes due to the diversity of the new task.  However, without distillation, the performance on the old classes is very poor.  After applying distillation and PPAS, the performance on the old classes is improved significantly (from 12.3\% to 63.5\%), while the mAP of the new classes remains as good (60.0\% vs 60.4\%).  Similarly, we equally split the 80 classes of COCO based on the category id.  Results on the ``40+40'' setup are shown in Table~\ref{tab:coco}.  We observe a consistent performance improvement compared to Seq-Fast in both experiment settings.  The effect of the size of $C_{\text{new}}$ will be further explored later.

\begin{table}[!tbh]
\small
\centering
\caption{Test results of ``40+40'' on COCO.  The reported performances are the mAP over all 80 classes.}
\label{tab:coco}
\begin{tabular}{l|cc}
 & \multicolumn{1}{l}{mAP @ 0.5} & \multicolumn{1}{l}{mAP @ [0.5 : 0.95]} \\ \hline \hline
Seq-Fast \shortcite{shmelkov2017incremental} & 37.4 & 21.3 \\
Ours w/ PPAS & 36.8 & 22.7 \\ \hline
$\mathbf{T}$(1-80) & 42.4 & 26.4
\end{tabular}
\end{table}

\begin{figure*}[!tbh]
    \centering
    \includegraphics[trim=0cm 1cm 0cm 0cm, width=1.\textwidth]{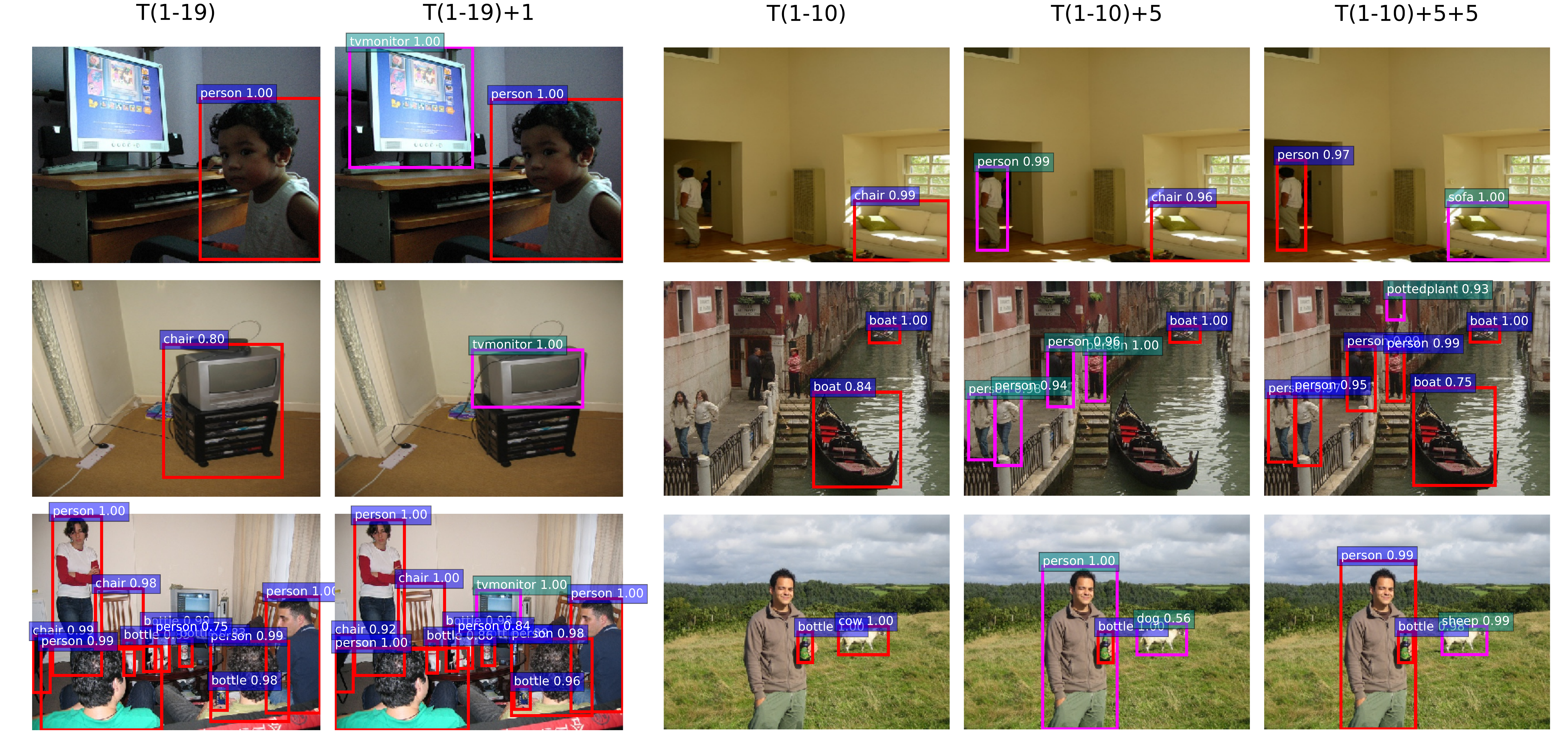}
    \caption{Examples of our lifelong detection results. The two columns on the left are the results of ``19+1'', and the three columns on the right depict the results of ``10+5+5''.  The old classes are plotted in red boxes, while the new classes are illustrated as pink boxes. Images are resized for better visualization.}
    \label{fig:sample}
    \vspace{-2mm}
\end{figure*}

\subsection{Adding classes sequentially}\label{sec:sequential}

In this subsection, we consider that multiple new tasks arrive sequentially.  Specifically, we generate three settings denoted as ``10+1...'', ``10+2...'', and ``10+5+5''.  All experiments are based on $\mathbf{T}$(1-10) trained on the first 10 classes of VOC dataset as $C_{\text{old}}$.  The difference is how the rest of the 10 classes are given.  For the case of ``10+1...'',  the last 10 classes are divided into 10 sequential tasks, and each task consists of a single object class.  The other two settings are constructed in a similar fashion.  It is worth pointing out that the number of training iterations for each task is set to be 4K/8K/20K if 1/2/5 classes are added.  This ensures the total number of iterations for updating all 10 new classes is the same for three sequential experiments.   
We run the experiment five times and
report the final mAP (after all 20 classes have been seen)  in Table~\ref{tab:sequential}.

Table~\ref{tab:sequential} compares our method to other baselines.  When the size of the new task is small ($|C_{\text{new}}|$=1,2), the mAP of our method is on par with Seq-Fast.  When the size of the new task gets larger, our method outperforms Seq-Fast (56.7\% vs 55.2\% for $|C_{\text{new}}|$=5, and 61.8\% vs 60.8\% for $|C_{\text{new}}|$=10).   
For \textbf{ablation} study, our method achieves much higher mAP in all four sequential learning settings compared to the fine-tuning baseline and distillation without PPAS, demonstrating the effectiveness of our method.


Figure~\ref{fig:sample} showcases the detection results after adding new classes, illustrating the ability of sequentially extending object detectors to new classes without forgetting the old classes.



\begin{table}[!tbh]
\small
\centering
\caption{Comparison of the inference time.  Running time is measured by averaging the inference time of $\sim$5K VOC test images on a single K80 GPU.}
\label{tab:time}
\begin{tabular}{l|cccc}
& \# proposals  &  total time & rate & gain  \\
\hline \hline
Seq-Fast \shortcite{shmelkov2017incremental} & 2,000 & 1.91s & 0.5 fps & 1x \\
Ours & 300 & 0.34s & 3 fps & 6x \\
\end{tabular}
\end{table}

\subsection{Comparison of running time}
One of the advantages of our proposed lifelong detection compared to Seq-Fast is that our method dynamically adapts the proposal network and provides a much faster inference.   To quantify the improvement, we compare the running time of the two methods in Table~\ref{tab:time}.  For Seq-Fast, to ensure a good recall on object proposals, a large number of proposals needs to be processed \cite{ren2015faster}, which deteriorates the detection speed.  Particularly, the external proposal computation by Edgebox alone takes 0.25s \cite{zitnick2014edge}, and the network inference requires another 1.66s to extract convolutional features and classify 2,000 RoIs, leading to a frame rate of 0.5 fps.  In contrast, being built around the Faster R-CNN architecture, our approach achieves recalls on par with Edgebox but with much fewer proposals.   In addition, there is also less overhead since the RPN and the R-CNN share the convolutional features.  Together, our method reaches a frame rate of 3 fps, which is a \textbf{6x} improvement compared to Seq-Fast.

\section{Conclusion}
In this paper, we study the problem of lifelong learning in object detection, where the data of new object classes arrive sequentially.  To overcome catastrophic forgetting, we propose to employ the teacher-student scheme to apply knowledge distillation on both RPN and R-CNN of a Faster R-CNN detector.  Furthermore, a pseudo-positive-aware sampling strategy is presented, which prefers valuable candidates in sample selection for effective distillation.  Evaluations on PASCAL VOC 2007 and COCO datasets demonstrate that our method compares favorably to existing baselines in both detection accuracy and inference speed. 


\FloatBarrier
\clearpage

{\small
\bibliographystyle{aaai}
\bibliography{ref}
}
\clearpage
\appendix
\clearpage
\section{Implementation Details}
\label{app:imp_detail}
We use SGD with momentum as the optimizer, and process a single image in each iteration.  When training network $\mathbf{T}$, our learning rate scheduler works as follows.  The learning rate is initially set to 0.001 for the first $k$ iterations and then annealed to 0.0001, where $k$ is 50K and 350K for VOC and COCO, respectively.   The total number of training iterations used for VOC and COCO are 70K and 490K.   We set the momentum of SGD to 0.9 and use a weight decay of 0.0001.  When training $\mathbf{S}$ for a new task, the learning rate of 0.0001 is used.  If the new task contains a single class, we train $\mathbf{S}$ for 4K iterations.  Otherwise, we set the number of training iterations to the same number as the one used for training  $\mathbf{T}$.  

\section{Varying the new class in ``19+1'' experiment}
\label{app:vary_19+1}

\begin{table}[!htbp]
\small
\centering
\caption{Test results of all possible ``19+1'' experiments on VOC 2007. Here the new class is enumerated (denoted with ``+'' in front), and the rest of the 19 classes are the old classes. The reported mAP is based on all 20 classes.}
\label{app:vary_new}
\begin{tabular}{l|ccccc}

\hline\hline
New & +plane & +bicycle & +bird & +boat & +bottle  \\
\hline
mAP & 68.1 & 69.4 & 68.8 & 67.8 & 69.0  \\
\hline\hline
New & +bus & +car & +cat & +chair & +cow  \\
\hline
mAP & 69.0 & 70.7 & 69.1 & 67.2 & 66.7  \\
\hline\hline
New & +table & +dog & +horse & +mbike & +person  \\
\hline
mAP & 68.2 & 67.6 & 68.7 & 69.7 & 69.0  \\
\hline\hline
New & +plant & +sheep & +sofa & +train & +tv  \\
\hline
mAP & 69.8 & 67.8 & 68.9 & 69.1 & 69.7  \\
\hline\hline
\end{tabular}
\end{table}

To check the reproducibility of our method, we repeat the ``19+1'' experiment by taking each of the 20 classes in VOC 2007 as the new class and the rest of the 19 classes as the old classes. We train $\mathbf{T}$(1-19) with these 20 different versions of $C_{\text{old}}$, and subsequently add the corresponding new class. Table~\ref{app:vary_new} lists the mAPs for each experiment setting. The method performs consistently when given different new classes. The mAP ranges from 66.7\% for ``cow'' and 70.7\% for ``car'', with an average mAP for all test classes of 68.7\% and standard deviation of 0.9\%.

\section{Varying the number of classes in each incremental task}

\begin{figure*}[!tbhp]
    \centering
    \includegraphics[width=0.9\textwidth]{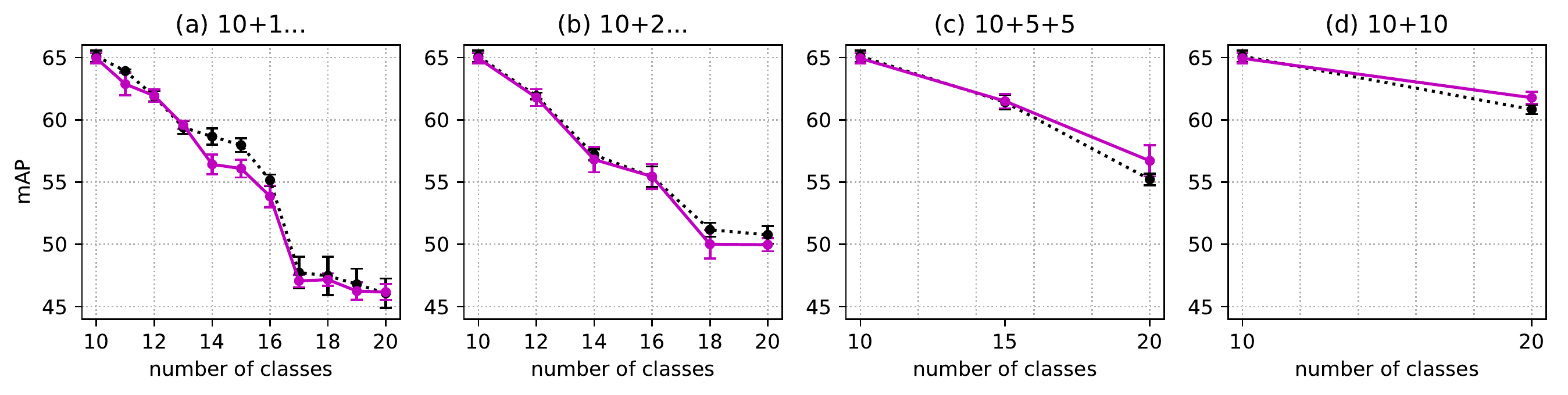}
    \caption{The mAP vs the number of classes the models have seen so far.  All experiments use the first 10 classes of VOC dataset as $C_{\text{old}}$ and the rest of the 10 classes are added sequentially.  In all four settings, the solid lines represent our method, and the dotted lines are Seq-Fast.}    
    \label{fig:vary_cls_size}
\end{figure*}

From Figure~\ref{fig:vary_cls_size}, we observe that as new classes are added, the overall mAP drops, which is caused by performance degradation on the old classes.  However, as the number of object classes per task increases, the final mAP of 20 classes is higher.  The improvement benefits from two factors. First, the larger size of a task in each sequential training step leads to better performance on the new classes.  This is because each class sees more variety of ``negative'' samples during training and thus predicts fewer false positives at testing.  Second, the larger size of a task often means the training images are more diverse.  Thus, it is more likely for these images to contain regions which help to better retain the performance on the old classes.

\section{Addition of similar classes}

\begin{table}[!tbhp]
\small
\centering
\caption{Effect of filtering out FP detections in $\mathbf{T}$ when sampling RoIs for distillation. The results are shown in the format of ``Old / New / All''.}
\label{app:filterByGT}
\begin{tabular}{l|c c c c}
          & Filter FP & w/o Filter FP \\
\hline \hline
``10+1...'' & 51.1 / 37.5 / 44.3 & 54.2 / 38.2 / 46.2  \\
``10+10'' & 63.2 / 60.6 / 61.9 & 63.5 / 60.0 / 61.8   \\
\end{tabular}
\end{table}

When a newly added class is visually similar to some seen classes of the teacher network (\emph{e.g.}, adding ``sheep'' to a network that is able to detect ``cow''),  it is likely that $\mathbf{T}$ would mistakenly detect ``sheep'' in the new training image as ``cow'' initially.   While the model needs to correct the prediction by lowering the prediction of the old classes, these misclassified regions (\emph{i.e.}, false positive ones) carry meaningful knowledge of the old classes and therefore they are valuable for distillation. We conduct experiments to see how the performance changes if the false positive (FP) detections in $\mathbf{T}$ are excluded from distillation samples.  

Before sampling the 128 RoIs for distillation using PPAS strategy, we first filter out RoIs that have IoU greater than 0.5 with ground-truth boxes of the new classes from $\mathbf{T}$.  We denote this setting as ``Filter FP'',  which is compared with its counterpart without filtering in Table~\ref{app:filterByGT}.  For the ``10+1...'' experiment, the ``Filter FP'' performs worse compared to the method without filtering those meaningful RoIs for distillation. On the other hand, if many classes are added together (``10+10'’), the mAP is not harmed.  This is possibly due to the size and variety of the training images in $C_{\text{new}}$ which improves the effectiveness of the distillation.  It is worth mentioning that the biggest drop in performance in Figure~\ref{fig:vary_cls_size}(a) happens when adding ``sheep'' (the 17th class in VOC) to the detector.  The performances on old classes ``cow'', ``dog'', and ``horse'' are affected after adding the new class, and thus, the overall mAP drops.  Finding a means to better maintain the performance on old classes when new classes are extremely similar to the old ones are considered as a part of the future work.

\section{Study on components of PPAS}

\begin{table}[!htbp]
\small
\centering
\caption{Ablation studies of each component of PPAS. The experiments are conducted in the``10+1...'' setting.}
\label{app:10+1}
\begin{tabular}{l|l|ccc}
\multicolumn{2}{l|}{} & Old & New & All \\ \hline \hline
\multicolumn{2}{l|}{$\mathbf{T}$(1-10)} & 64.9 & - & - \\ \hline
\rotatebox{90}{FT} & Entire & \multicolumn{1}{c}{3.2} & \multicolumn{1}{c}{2.1} & \multicolumn{1}{c}{2.7} \\ \hline
\multirow{6}{*}{\rotatebox{90}{Distil}} & Ours w/o PPAS & 22.9 & 17.7 & 20.3 \\
& Ours w/o RPN Filtering & 54.1 & 38.1 & 46.1 \\
& Ours w/o RPN Top-score & 53.6	& 38.0	& 45.8	 \\
& Ours w/o R-CNN Filtering & 46.7	& 31.6	& 39.2 \\
& Ours w/o R-CNN Top-score & 32.9	& 20.2	& 26.6 \\
& Ours w/ PPAS & 54.2 & 38.2 & 46.2 \\ \hline
\multicolumn{2}{l|}{$\mathbf{T}$(1-20)} & \multicolumn{1}{l}{69.9} & \multicolumn{1}{l}{70.5} & \multicolumn{1}{l}{70.2}
\end{tabular}
\end{table}

We analyze the effectiveness of each component of PPAS and demonstrate their impact on the final performance.  All of the ablation studies here are performed under the ``10+1...'' setting.   We exclude each of the four components of PPAS and run the experiment to see how the performance changes, and report the results in Table~\ref{app:10+1}.   For RPN, if we do not exclude pseudo-positive anchors from being selected as negative anchors (``Ours w/o RPN Filtering''), the mAP of all classes drops by 0.1\%.   And if we sample the anchors randomly instead of preferring anchors with high objectness scores (``Ours w/o RPN Top-score''), the mAP drops by 0.4\%.   For R-CNN, if the pseudo-positive RoIs are not excluded from being treated as negative RoI samples (``Ours w/o R-CNN Filtering''), the mAP is slashed by 7.0\%, while if the RoIs used for distillation are sampled randomly (``Ours w/o R-CNN Top-score''), the mAP is largely reduced by 19.6\%. 
Selecting RoIs with high non-background scores instead of random sampling for R-CNN distillation affects the performance the most. Excluding pseudo-positive RoIs from negative RoI samples also largely boosts the mAP.  On the other hand, the mAP drop caused by the RPN is possibly compensated by the R-CNN since the R-CNN is updated accordingly trying to overcome the degradation in proposal generation.

\end{document}